\documentclass[conference]{IEEEtran}
\IEEEoverridecommandlockouts
\usepackage{amsmath,graphicx}
\usepackage{graphicx}
\usepackage{subfigure}
\usepackage{stfloats}
\usepackage{caption}
\usepackage{algorithmic}
\usepackage{algorithm}
\usepackage{amssymb,amsfonts}
\usepackage{color}
\usepackage{float}
\usepackage{threeparttable}
\usepackage{soul}
\usepackage{booktabs}
\usepackage{multirow}
\usepackage{booktabs}
\usepackage{siunitx}
\usepackage{booktabs}
\usepackage{siunitx}
\usepackage{adjustbox}
\usepackage{lipsum} % http://ctan.org/pkg/lipsum
\usepackage{cite}
\usepackage{amsmath,amssymb,amsfonts}
\usepackage{algorithmic}
\usepackage{graphicx}
\usepackage{textcomp}

\def\BibTeX{{\rm B\kern-.05em{\sc i\kern-.025em b}\kern-.08em
    T\kern-.1667em\lower.7ex\hbox{E}\kern-.125emX}}
\begin{document}

\title{Novel Video Prediction for Large-scale Scene using Optical Flow}

\author{\IEEEauthorblockN{Henglai Wei}
\IEEEauthorblockA{\textit{University of Victoria}\\
Victoria, Canada \\
henglaiwei@uvic.ca}
\and
\IEEEauthorblockN{Xiaochuan Yin}
\IEEEauthorblockA{\textit{Tongji University}\\
Shanghai, China \\
yinxiaochuan@hotmail.com}
\and
\IEEEauthorblockN{Penghong Lin}
\IEEEauthorblockA{\textit{Horizon Robotics Inc.}\\
Beijing, China \\
penghong.lin@hobot.cc}
%\and
%\IEEEauthorblockN{Yang Shi}
%\IEEEauthorblockA{\textit{University of Victoria}\\
%Victoria, Canada \\
%yshi@uvic.ca}
}

\maketitle
\begin{abstract}
 Making predictions of future frames is a critical challenge in autonomous driving research. Most of the existing methods for video prediction attempt to generate future frames in simple and fixed scenes. In this paper, we propose a novel and effective optical flow conditioned method for the task of video prediction with an application to complex urban scenes. In contrast with previous work, the prediction model only requires video sequences and optical flow sequences for training and testing. Our method uses the rich spatial-temporal features in video sequences. The method takes advantage of the motion information extracting from optical flow maps between neighbor images as well as previous images. Empirical evaluations on the KITTI dataset and the Cityscapes dataset demonstrate the effectiveness of our method. %Experiments demonstrate that our method significantly outperforms prior methods on prediction across diverse and complex visual scenes.
\end{abstract}
\section{Introduction}
\label{sec:intro}
%Autonomous driving is currently emerging
Humans are remarkably capable of predicting object motions and scene dynamics over short timescales. But why do humans excel at this work? Generally, we depend on extensive knowledge about the real-world through our past visual experience that consisted of rich objects and interactive relationships of different scenes to make future predictions. We can apply accumulated knowledge when predicting a new scene for a short time, even a little longer period. The prediction ability is also important for the intelligent agents and autonomous systems, because it is a useful and valuable auxiliary method for the task of path planning and decision making. To accomplish the prediction task, we need a model that is usually required to understand the scenes and how the scenes may transform. It is a challenging task for the autonomous vehicle, because the diverse objects and physics rules of the visual scenes are difficult to describe.
\begin{figure}
\centering
\includegraphics[width=\columnwidth]{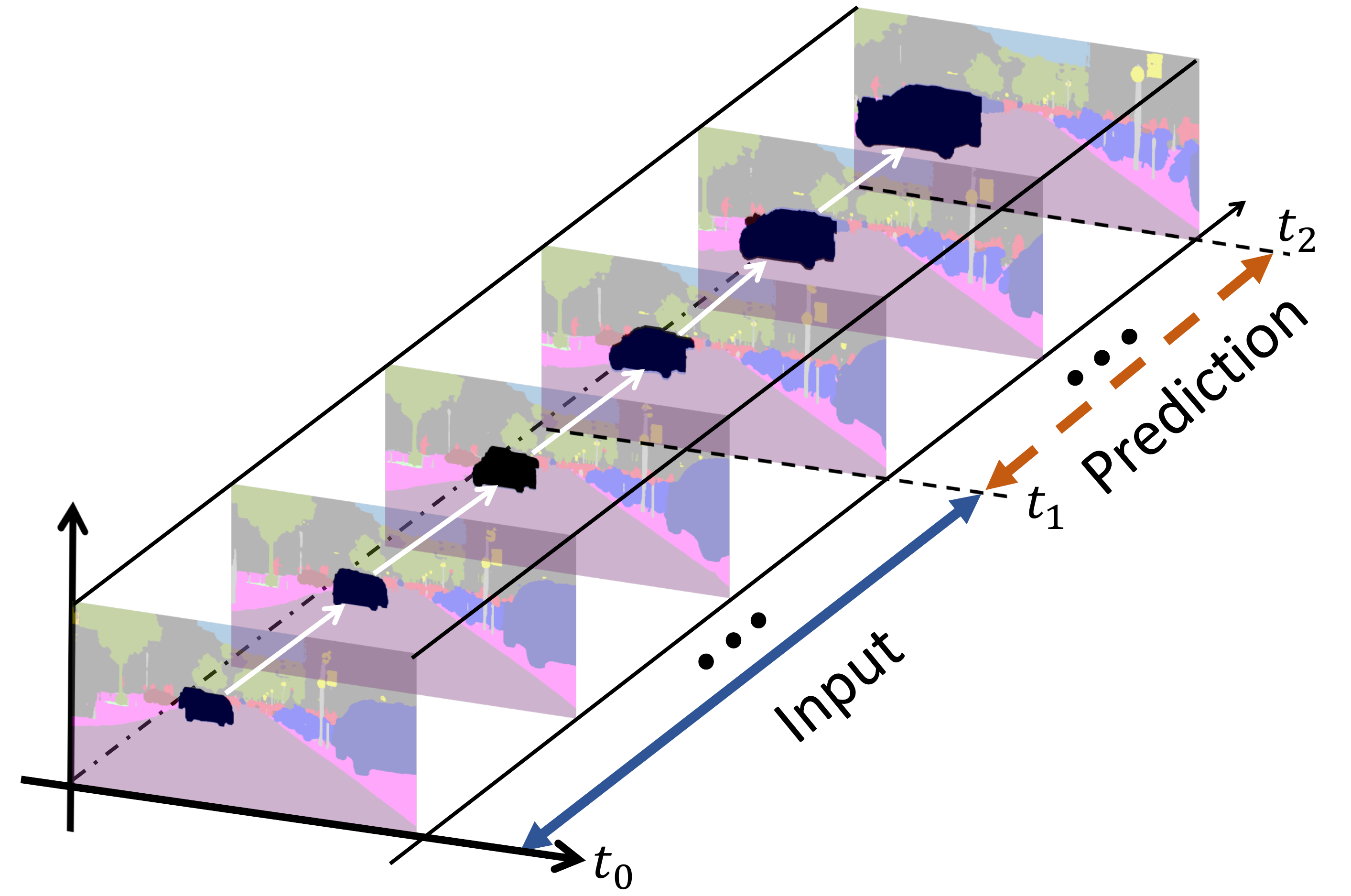}
\caption{Illustration of our purpose. As shown, previous frames from $t_0$ to $t_1$ represent the input of the model. The frames from $t_1$ to $t_2$ are the prediction of future frames. Each frame contains the spatial information, and multiple frames as inputs contain the temporal information. All the frames come from the Cityscapes dataset. Black part in the six frames is a moving vehicle.}
\label{fig:figure1}
\end{figure}

We are particularly inspired by recent work that made video prediction about the motion of human and the manipulator in an end-to-end framework \cite{finn2016unsupervised}. In their paper, the background are fixed and the interaction between different objects are simple. In addition, limited number of explicit states and motions are difficult to deal with the complex urban driving scenes. Therefore, in this paper, we introduce the optical flow to our framework because it implicitly contains valuable motion information of each pixel. we propose a novel predictive model that exploit spatial-temporal appearance information of previous frames and the inter-frame optical flow information as shown in Figure \ref{fig:figure1}. Our model has two independent input steams, one is for the previous RGB frames and the other is for the optical flow estimation. We take an end-to-end framework in allowing the model to map directly from input pixels to the prediction of next frame or longer. Our model can be trained simply using sequences of RGB images and optical flow maps without manual labeled data or camera ego-motion information. 

Our approach builds upon the insight that merge appearance of images and the motion of pixels extracted from the optical flow. The prediction model includes three core modules. First, we refer to as optical flow prediction networks (OFPN), which inputs 4 successive optical flow maps and outputs the prediction optical flow map. All the 5 optical flow maps then enter into the second part that is the motion estimation networks (MEN). The MEN is responsible for the motion estimation, and outputs dense transformation maps T. The last module, which we call spatial-temporal prediction networks (STPN), outputs the next frame or longer prediction frames. In \cite{finn2016unsupervised}, the authors developed an action-conditioned prediction model that explicitly models pixel motion. While limited number of explicit motion could produce some reasonable video prediction for certain types of scenes (e.g., static background scene), the same model would fail miserably when presented with another set of scenes with more diverse appearance and complex interaction between objects. Thus, we aim to formulate the optical flow motion estimation as the supplementary information for the video prediction neural networks. Experiments demonstrate that our method significantly outperforms prior methods on prediction across diverse and complex visual scenes. 

The main contributions of our paper are summarized as follows:
\begin{enumerate}
\item We propose a novel video prediction model that uses the previous frames and optical flow maps between neighbor frames.  
\item Our video prediction model achieves state-of-the-art performance on the KITTI dataset and the Cityscapes dataset. 
\item Our model is capable of making prediction of natural images and semantic segmentations.
\item Our model can learn jointly from the optical flow prediction loss and the frame prediction loss.
\end{enumerate}

The structure of this paper is organized as follows. In section 2, we introduce the related works about video prediction problem. In section 3, we introduce our framework for video prediction using optical flow. Experiments of video prediction using optical flow maps are reported and analyzed in section 4. We conclude our paper in the last section.

\section{Related work}
In this section, we survey the most related works. We first provide a literature review of video prediction with a particular focus on the video prediction using deep learning methods. %In the second part, we review recent works that focuses on optical flow estimation.

\subsection{Video prediction}
\textbf{Scenes of Video Prediction :} Motivated by the great success of deep learning in machine vision (e.g., image classification and object recognition) \cite{krizhevsky2012imagenet}.  Recently some models using the deep learning approaches have been proposed to tackle the problem of video prediction under different scenes. Early work about video prediction focused on low resolution video clips or images containing simple predictable content without any background such as the moving digit \cite{srivastava2015unsupervised,xue2016visual} and Atari Game prediction \cite{oh2015action}. Higher resolution natural scenes with static background are more complicated but promising. \cite{mathieu2015deep} proposed different models to predict the actions, poses or paths of human. \cite{finn2016unsupervised} built a model to make prediction of the robot arms. Generally these training images have static background, and visual representations are not that complex. There are some real-world videos that contain not only the moving targets but also the dynamic background such as the urban traffic scenes. \cite{lotter2016deep} predicted realistic looking frames, and \cite{neverova2017predicting} only predicted future semantic segmentations rather than natural frames. In contrast to these work, our model can predict complicated real-world scenes (e.g., urban scenes). At the same time, we also verify our model to predict the semantic segmentations.

\textbf{Methods of Video Prediction :} There have been a number of promising approaches for the task of video prediction. \cite{ranzato2014video} introduced a generative model that used the recurrent neural network (RNN) to predict the next frame. \cite{srivastava2015unsupervised} adapted a LSTM model for video prediction. \cite{mathieu2015deep} achieved sharper video prediction by using multi-scale architecture and an adversarial loss function. Rather than just transforming the pixels from previous frames, our method warps the optical flow features containing rich motion information over RGB images. \cite{luo2017unsupervised} presented an unsupervised representation learning approach to predict long-term 3D motions.  \cite{vondrick2016generating} proposed a generative adversarial network (GAN) for video prediction with a convolutional architecture that untangles the foreground of scenes from the background. \cite{finn2016unsupervised} developed an action-conditioned video prediction model by predicting a distribution over pixels. In a similar spirit, \cite{DBLP:journals/corr/WattersTWPBZ17} introduced a Convolution neural networks (CNN) for learning the dynamics of a physical system from raw visual observations. This work highlighted the importance of using the dynamics of the system. In contrast to the traditional deterministic method, \cite{xue2016visual} proposed a probabilistic method for the frame prediction to solve the uncertainty problem.
%We are hoping that our model are more general to predict the most complicated scenes. 
%introduce video prediction work from 'the method' view. I think after this stage , the part2 is almost finished.
\begin{figure*}[ht]
\centering
\includegraphics[width=2\columnwidth]{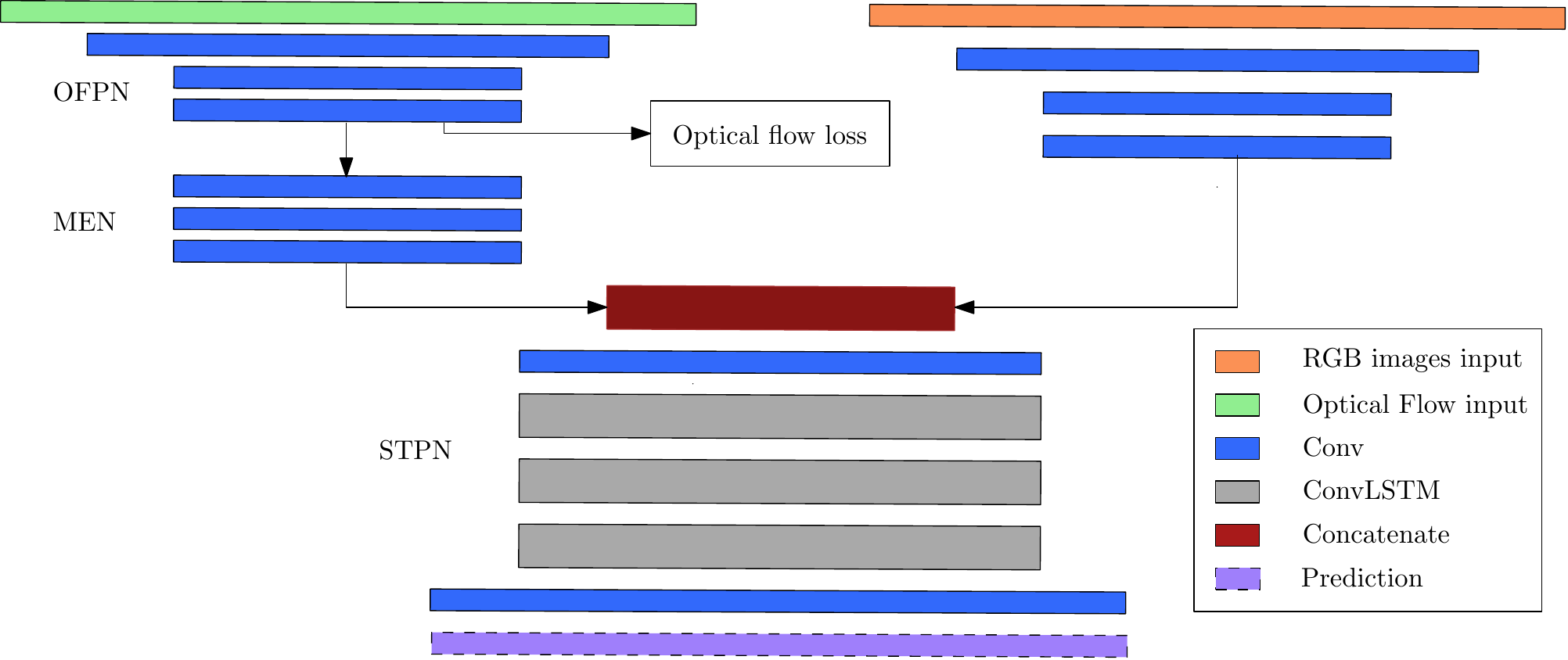}
\caption{\textbf{Video Prediction Network:}  We illustrate our network architecture for the video prediction (see the legend on the bottom right). The height of each block represents the spatial dimension of the feature maps and the output channels respectively, and the increase/reduction factor is 2. There are two separate input streams: a pathway of successive RGB frames, and a pathway of optical flow maps implicitly consisting the pixel motion information. In particular, we set an auxiliary loss for the future optical flow prediction in training. The outputs of two steams are fed into the spatial-temporal predictive networks to produce the RGB image or semantic image for the next time step.}
\label{fig:figure2}
\end{figure*}

\section{Approach}

In this section, we introduce our framework for video prediction as shown in Figure \ref{fig:figure2}. Our ultimate goal is to make accurate prediction of future frames in complex and real-world scenes. Given RGB frames observed at time steps $t_0,t_1,\dots,t_n$ and corresponding optical flow maps extracted from neighbor frames, our model takes advantage of the history information of frames to predict the future frames. In addition, our model is able to jointly train optical flow prediction and video prediction. 
%Here we first describe our overall approach for video prediction on large-scale unlabeled autonomous driving dataset. Next three core parts of our model are proposed. And then we will give the detailed architecture specifications.

%training a motion prediction CNN ,a pose estimation CNN from optical flow and a video prediction network from unlabeled video sequences. Our model computes the next frame by utlizing the motion information which is extracted from optical flow.

%by utilizing past RGB frames and the pixels' motion extracting from optical flow
%video prediction

The core modules of our model are the optical flow prediction networks (OFPN), the motion estimation networks (MEN) and the spatial-temporal prediction networks (STPN). Formally, let $\hat{\textbf{I}}=\{\hat{I}_1,\dots,\hat{I}_N\}$ be a sequence of predicted frames. Input RGB frames of video sequences and optical flow are denoted as $\textbf{I}=\{I_1,\dots,I_N\}$ and $\textbf{I}'=\{I'_1,\dots,I'_{N-1}\}$ respectively.

\subsection{Optical flow prediction networks (OFPN) } 

\begin{figure}[h]
\centering
\subfigure[Extracting optical flow map from the neighbor frames.]{
\label{fig:of_a} %% label for first subfigure
\includegraphics[width=\columnwidth]{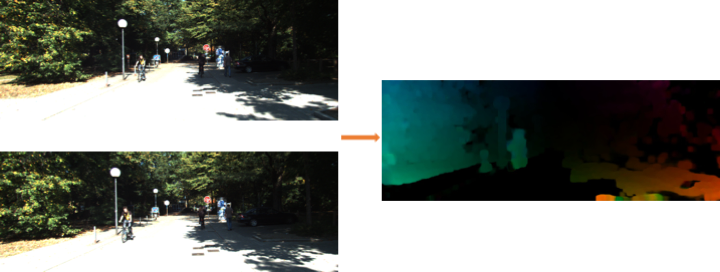}}\hfill
%\hspace{1in}
\subfigure[Illustration of the optical flow. The point $p_t$ move from $(x,y)$ to the next location $(x+\Delta x,y+\Delta y)$ at $t+1$.]{
\label{fig:of_b} %% label for second subfigure
\includegraphics[width=\columnwidth]{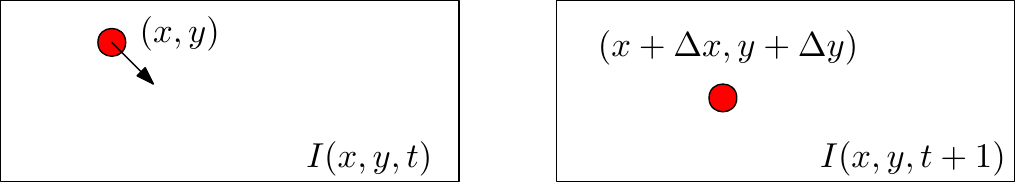}}
\caption{Given a reference frame $I_t$ and its neighbor frame $I_{t+1}$, the optical flow field $M_{t\to t+1}=\mathcal{F}(I(x,y,t),I(x,y,t+1))$ is estimated.}
% \protect \cite{farneback2003two}.}
\label{fig:optical flow} %% label for entire figure
\end{figure}

Optical flow is a vector field, having the same size as the RGB frames, with 2D flow vector per pixel \cite{horn1981determining}. Optical flow represents the apparent displacements of pixels in $x$ and $y$ directions due to the relative motion between consecutive frames see Figure \ref{fig:of_b}. Acquiring optical flow field needs precise per-pixel localization, and also requires finding correspondences between a pair of input RGB frames (e.g.,$I_t$ and $I_{t+1}$) shown in Figure \ref{fig:of_a}. 

As we want to predict the next RGB frame, the last optical flow that extracts between the predicted frame and the frame before that is unknown. We therefore use the OFPN to predict the future optical flow. In this approach, the OFPN is trained as an auxiliary task to predict the optical flow $\hat{I}'_N$ over the previous optical flow $\textbf{I}'=\{I'_1,\dots,I'_{N-1}\}$. By this way, we can get the optical flow frame $\textbf{I}'=\{I'_1,\dots,I'_{N-1},\hat {I}'_N\}$. We can formulate the objective function for the optical flow prediction as:

\begin{equation} \label{eq:1}
\ell_{of}=||\hat{I}'_{N}-I'_{N}||_2^2%\sum_{I'_1,\dots,I’_{N-1}}
\end{equation}
where $\ell_{of}$ is an optical flow prediction loss.

Minimizing the above loss equation \ref{eq:1} can generate the predicted optical flow field $\hat{I}'_{N}$, which perfectly minimize the auxiliary loss. The predicted optical flow together with previous optical flow maps are applied as input to the motion estimation networks. 

\subsection{Motion estimation networks (MEN)}

The motion estimation networks (MEN) produces a transformation feature map $T$, representing the dense motion information of pixels (i.e. the rotation and displacement of every pixel). The optical flow implicitly consists of the object motion. Unlike \cite{finn2016unsupervised}, we utilize the optical flow to represent the motion field rather that the limited number of states and actions in our approach.

The MEN uses the 3D convolutional layers with kernel ($3\times 3\times 3$) to generate the motion transformation for each pixel. 

Note that the inputs to the MEN are the predict optical flow frame $\hat I'_N$ from OFPN and previous optical flow frames $\textbf{I}'$, and the outputs are the transformation map $T$ that consists of relative displacement and rotation between the consecutive frames. Therefore, the transformation maps $T$  are used as input to the spatial-temporal prediction networks (STPN), wrapping the corresponding RGB frames to predict the future frame. 

\subsection{Spatial-temporal prediction networks (STPN)} 

The STPN is one core part of the proposed model. It can generate the prediction frame. To generate more accurate prediction frame, we employ stacked Convolutional LSTM layers (ConvLSTM) \cite{xingjian2015convolutional} in the STPN. Recurrence through convolutions is good at multi-step video prediction because it utilizes the spatial invariance of image, as well as the temporal information between the neighbor frames. 

More formally, the transformation $T$ applied to the current frame $I_t$ produces the next predict frame $\hat I_{t+1}$. Let $(x_t,y_t)$ denote the homogeneous coordinates of pixels in the current image and $(\hat x_{t+1}, \hat y_{t+1})$ in the next image.

\begin{equation}
\left(                 
  \begin{array}{c}  
    \hat x_{t+1}\\  
    \hat y_{t+1}   
  \end{array}\right)=T\left(\begin{array}{c}  
    x_{t}\\  
    y_{t}\\
    1   
  \end{array}
\right) 
\end{equation}

Training the STPN comes down to minimizing the prediction error between the predict frame $\hat I_N$ and the ground truth frame $I_N$. Inspired by \cite{mathieu2015deep}, the $\ell_1+\ell_{gdl}$ loss is used in the STPN, 

\begin{equation}
\ell_{st}(\hat{I}_N,I_N) = \ell_1(\hat{I}_N,I_N)+\ell_{gdl}(\hat{I}_N,I_N)
\end{equation} 
where $\ell_{gdl}$ is the gradient difference loss, defined as
\begin{equation}
\begin{split}
\ell_{gdl}(\hat{I},I) &=\sum_{i,j}\left||I_{i,j}-I_{i-1,j}|-|\hat{I}_{i,j}-\hat{I}_{i-1,j}|\right|^\alpha \\
&+ \sum_{i,j}\left||I_{i,j-1}-I_{i,j}|-|\hat{I}_{i,j-1}-\hat{I}_{i,j}|\right|^\alpha
\end{split}
\end{equation}
where $\alpha=1$, and $|\cdot|$ denotes the absolute value function. $\hat{I}$ and $I$ denote $\hat{I}_N$ and $I_N$ respectively.

Our final objective becomes,
\begin{equation}
\ell_{final}=\lambda_{of}\ell_{of}+\lambda_{st}\ell_{st}
\end{equation}
where $\lambda_{of}$ and $\lambda_{st}$ are the weighting for the optical flow loss and prediction networks respectively. When training the model we choose $\lambda_{of}=\lambda_{st}=1$. 

\subsection{Network architecture}

For the optical flow prediction network (OFPN), we use 3 convolutional layers and all layers are followed by batch normalization (BN) and ReLU activation. In the Motion estimation networks (MEN), 4 3D-convolutional layers are used to generate the transformation maps. We employ ConvLSTM \cite{xingjian2015convolutional} as backbone model of the spatial-temporal prediction networks. This architecture captures spatial-temporal correlations very well. The detailed specifications of the three parts are shown in Table \ref{tab:network architecture}.

\begin{table}%[h]
    \centering
    \begin{tabular}{l c c}
    \toprule
        Layer type & Kernel size & Feature maps\\
        \midrule
        \multicolumn{3}{c}{Optical flow prediction networks (OFPN)}\\
        \midrule
         2D-Conv-BN-ReLU& (3,3) &32\\
         2D-Conv-BN-ReLU& (3,3) &64\\
         2D-Conv-BN-ReLU& (3,3) &128\\
         \midrule
         \multicolumn{3}{c}{Motion estimation networks (MEN)}\\
    \midrule
        %Layer type & Kernel size & Channels\\
        %\midrule
         3D-Conv-BN-ReLU& (3,3,3) &64\\
         3D-Conv-BN-ReLU& (3,3,3) &64\\
         3D-Conv-BN-ReLU& (3,3,3) &64\\
         \midrule
         \multicolumn{3}{c}{Spatial-temporal prediction networks (STPN)}\\
    \midrule
        %Layer type & Kernel size & Channels\\
        %\midrule
         2D-ConvLSTM-BN-ReLU& (3,3) &128\\
         2D-ConvLSTM-BN-ReLU& (3,3) &96\\
         2D-ConvLSTM-BN-ReLU& (3,3) &64\\
         2D-ConvLSTM-BN-ReLU& (3,3) &32\\
         \bottomrule
    \end{tabular}
    \caption{Network architecture of our model}
    \label{tab:network architecture}
\end{table}

%%-----------------------------------------------

\begin{figure*}[htbp]
\centering
\subfigure[Input]{\label{fig:input_11}
\begin{minipage}[t]{0.23\textwidth}
\includegraphics[width=1.0\textwidth,height=0.35\columnwidth]{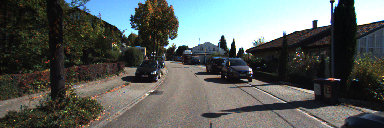}\vspace{0.5ex}
\includegraphics[width=1.0\textwidth,height=0.35\columnwidth]{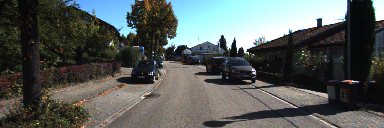}\vspace{0.5ex}
\includegraphics[width=1.0\textwidth,height=0.35\columnwidth]{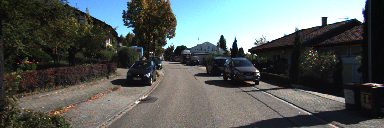}\vspace{0.5ex}
\includegraphics[width=1.0\textwidth,height=0.35\columnwidth]{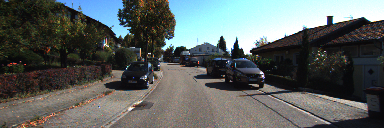}\vspace{0.5ex}
\includegraphics[width=1.0\textwidth,height=0.35\columnwidth]{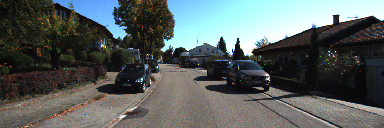}\vspace{0.5ex}
\end{minipage}
}
\subfigure[Ground Truth]{\label{fig:gt_11}
\begin{minipage}[t]{0.23\textwidth}
\includegraphics[width=1.0\textwidth,height=0.35\columnwidth]{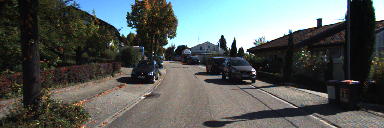}\vspace{0.5ex}
\includegraphics[width=1.0\textwidth,height=0.35\columnwidth]{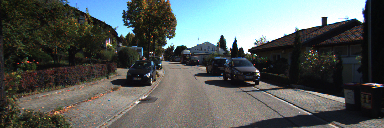}\vspace{0.5ex}
\includegraphics[width=1.0\textwidth,height=0.35\columnwidth]{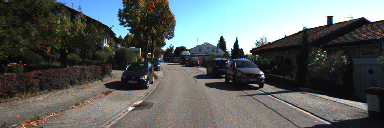}\vspace{0.5ex}
\includegraphics[width=1.0\textwidth,height=0.35\columnwidth]{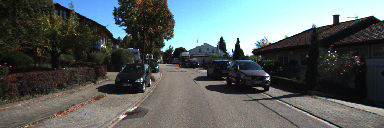}\vspace{0.5ex}
\includegraphics[width=1.0\textwidth,height=0.35\columnwidth]{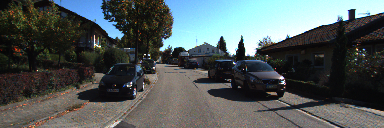}\vspace{0.5ex}
\end{minipage}
}
\subfigure[PredNet]{\label{fig:prednet_11}
\begin{minipage}[t]{0.23\textwidth}
\includegraphics[width=1.0\textwidth,height=0.35\columnwidth]{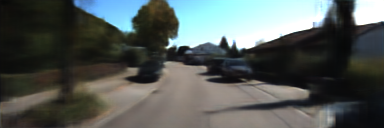}\vspace{0.5ex}
\includegraphics[width=1.0\textwidth,height=0.35\columnwidth]{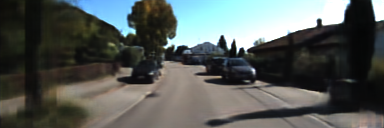}\vspace{0.5ex}
\includegraphics[width=1.0\textwidth,height=0.35\columnwidth]{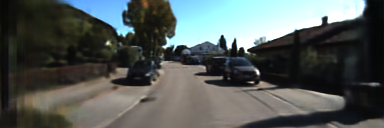}\vspace{0.5ex}
\includegraphics[width=1.0\textwidth,height=0.35\columnwidth]{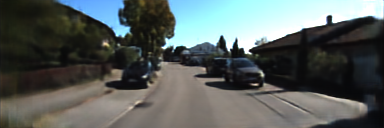}\vspace{0.5ex}
\includegraphics[width=1.0\textwidth,height=0.35\columnwidth]{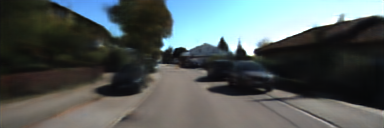}\vspace{0.5ex}
\end{minipage}
}
\subfigure[Our networks]{\label{fig:convlstm_11}
%\subfigure[Our networks]{\label{fig:our_11}
\begin{minipage}[t]{0.23\textwidth}
\includegraphics[width=1.0\textwidth,height=0.35\columnwidth]{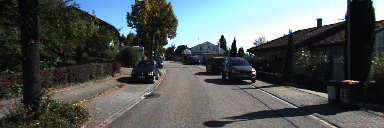}\vspace{0.5ex}
\includegraphics[width=1.0\textwidth,height=0.35\columnwidth]{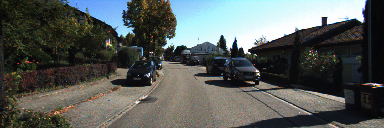}\vspace{0.5ex}
\includegraphics[width=1.0\textwidth,height=0.35\columnwidth]{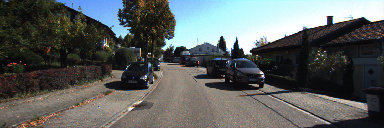}\vspace{0.5ex}
\includegraphics[width=1.0\textwidth,height=0.35\columnwidth]{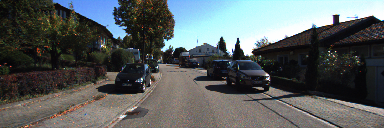}\vspace{0.5ex}
\includegraphics[width=1.0\textwidth,height=0.35\columnwidth]{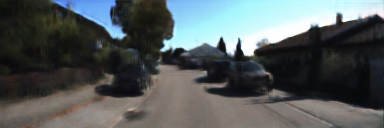}\vspace{0.5ex}
\end{minipage}
}
%\subfigure[Our networks]{\label{fig:our_11}
%\begin{minipage}[t]{0.181\textwidth}
%\includegraphics[width=1.0\textwidth,height=0.35\columnwidth]{pics/13_541-p-0.png}\vspace{0.5ex}
%\includegraphics[width=1.0\textwidth,height=0.35\columnwidth]{pics/13_541-p-1.png}\vspace{0.5ex}
%\includegraphics[width=1.0\textwidth,height=0.35\columnwidth]{pics/13_541-p-2.png}\vspace{0.5ex}
%\includegraphics[width=1.0\textwidth,height=0.35\columnwidth]{pics/13_541-p-3.png}\vspace{0.5ex}
%\includegraphics[width=1.0\textwidth,height=0.35\columnwidth]{pics/13_546p0.png}\vspace{0.5ex}
%\includegraphics[width=1.0\textwidth,height=0.35\columnwidth]{ls_minus_five_step_000525.png}\vspace{0.5ex}
%\end{minipage}
%}
\caption{Examples of video prediction results. For each frame, we show (a) the input frames, (b) ground truth frames, (c) results generated by PredNet method and (d) results produced by our networks trained with $\ell_1+\ell_{gdl}$ loss. Different rows show the results of different time steps. The input frames are at $t=1,2,3,4,5$. }
\label{fig:figure4}
\end{figure*}
% with 5 consecutive input frames from the KITTI dataset
\section{Experiments}

In this section, we evaluate the performance of our model and make comparison with previous approaches of video prediction. We evaluate our system on the Cityscapes dataset \cite{Cordts2016Cityscapes} as well as on the KITTI dataset \cite{Geiger2012CVPR}.

\textbf{Training details:} We implemented the model by using the Tensorflow and Keras \cite{chollet2015keras}. For all the experiments, we used batch normalization for all the layers except the output layers, and the Adam \cite{kingma2014adam} optimizer with the suggested super parameters (learning rate $lr=0.001, \beta_1=0.9, \beta_2=0.999$). Our systems are trained and deployed on a NVIDIA TITAN X GPU with 12GB memory. During training and testing stage, we resize the image sequences of the KITTI dataset to $128\times384$ and images of the Cityscapes dataset to $256\times512$. Our networks are trained in an end-to-end way.

\subsection{Video prediction of natural scenes}

In our first set of experiments, we train our system on real-world sequences from the KITTI dataset, with about 20K frames in total. No other data augmentation is used. The dataset has 18K and 2K images for training and testing respectively. Training and testing sequences are given in the form of tuple $(\textbf{I},\textbf{I}',\hat{I})$, where $\textbf{I}$ are the input raw RGB frames, $\textbf{I}'$ are the optical flow maps input and $\hat{I}$ is the predicted frame. We train our model using a frame rate of 1, and taking 5 consecutive RGB frames and 4 corresponding optical flow maps as the input. That is, we pass in initial RGB natural sequences i.e., $\textbf{I}=\{I_{t-4}, I_{t-3}, I_{t-2}, I_{t-1}, I_t\}$, as well as the corresponding optical flow maps include $\textbf{I}'=\{I'_{t-4}, I'_{t-3}, I'_{t-2}, I'_{t-1}\}$, then roll out the model sequentially, passing in the motion predictions from the previous optical flow maps. 
 
In Figure \ref{fig:figure4}, we compare the video prediction results using different approaches from the KITTI test set. In terms of complex real-world urban scenes, our results match the ground truth frames significantly better than previous video prediction methods. When compared with the ground truth, PredNet produces the blurry prediction results and does not give the accurate prediction as shown in Figure \ref{fig:prednet_11}. The results generated by PredNet is very similar to the input frames rather than the ground truth frames. Overall, the results of our approach tends to be more accurate and sharper. 

To evaluate the quality of the video prediction generating from different models, we compute the Peak Signal to Noise Ratio (PSNR) and the Structural Similarity Index Measures (SSIM) between the true frame $I_N$ and the predict frame $\hat{I}_N$ \cite{wang2004image}. The SSIM ranges from $-1$ to $1$, and larger score represents better prediction performance. In table 1, we presents the quantitative comparisons among different methods. We can clearly see the advantage of our method over other methods. The optical flow containing the motion features are fed into the STPN to yield the higher quality prediction. 

\begin{table}%[h]
\begin{center}
\begin{tabular}{l c c}
\toprule
Method  & PSNR & SSIM \\
\midrule
PredNet \cite{lotter2016deep}  & 15.0817 & 0.4738 \\
%ConvLSTM \cite{xingjian2015convolutional} &34.8492 &0.8722\\%&0.9022\\%& 0.0165 & 17.8098 \\
Our networks & \textbf{41.6849} & \textbf{0.9097}\\
\bottomrule
\end{tabular}
\end{center}
\caption{Mean pixel PSNR and SSIM on the KITTI dataset for video prediction with two input streams.}
\label{tab:natural performance}
\end{table}

\begin{figure*}[ht]
\centering
\subfigure[Input]{\label{fig:in_22}
\begin{minipage}[t]{0.23\textwidth}
\includegraphics[width=1.0\textwidth,height=0.35\columnwidth]{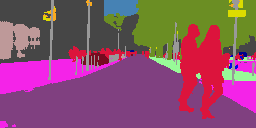}\vspace{0.5ex}
\includegraphics[width=1.0\textwidth,height=0.35\columnwidth]{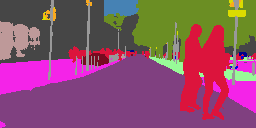}\vspace{0.5ex}
\includegraphics[width=1.0\textwidth,height=0.35\columnwidth]{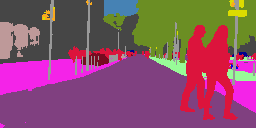}\vspace{0.5ex}
\includegraphics[width=1.0\textwidth,height=0.35\columnwidth]{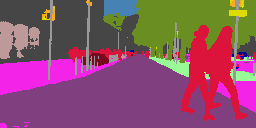}\vspace{0.5ex}
\includegraphics[width=1.0\textwidth,height=0.35\columnwidth]{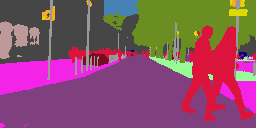}\vspace{0.5ex}
\end{minipage}
}
\subfigure[Ground Truth]{\label{fig:gt_22}
\begin{minipage}[t]{0.23\textwidth}
\includegraphics[width=1.0\textwidth,height=0.35\columnwidth]{66-r-1.png}\vspace{0.5ex}
\includegraphics[width=1.0\textwidth,height=0.35\columnwidth]{66-r-2.png}\vspace{0.5ex}
\includegraphics[width=1.0\textwidth,height=0.35\columnwidth]{66-r-3.png}\vspace{0.5ex}
\includegraphics[width=1.0\textwidth,height=0.35\columnwidth]{66-r-4.png}\vspace{0.5ex}
\includegraphics[width=1.0\textwidth,height=0.35\columnwidth]{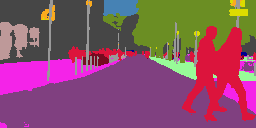}\vspace{0.5ex}
\end{minipage}
}
\subfigure[ConvLSTM]{\label{fig:convlstm_22}
\begin{minipage}[t]{0.23\textwidth}
\includegraphics[width=1.0\textwidth,height=0.35\columnwidth]{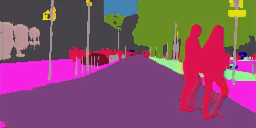}\vspace{0.5ex}
\includegraphics[width=1.0\textwidth,height=0.35\columnwidth]{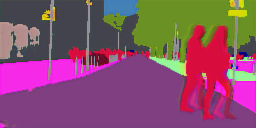}\vspace{0.5ex}
\includegraphics[width=1.0\textwidth,height=0.35\columnwidth]{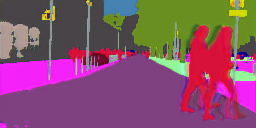}\vspace{0.5ex}
\includegraphics[width=1.0\textwidth,height=0.35\columnwidth]{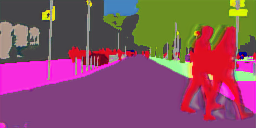}\vspace{0.5ex}
\includegraphics[width=1.0\textwidth,height=0.35\columnwidth]{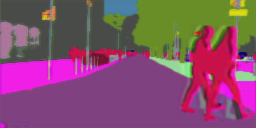}\vspace{0.5ex}
\end{minipage}
}
\subfigure[Our networks]{\label{fig:our network_22}
\begin{minipage}[t]{0.23\textwidth}
\includegraphics[width=1.0\textwidth,height=0.35\columnwidth]{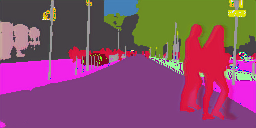}\vspace{0.5ex}
\includegraphics[width=1.0\textwidth,height=0.35\columnwidth]{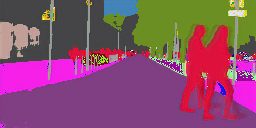}\vspace{0.5ex}
\includegraphics[width=1.0\textwidth,height=0.35\columnwidth]{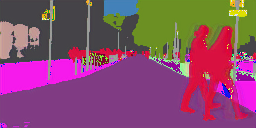}\vspace{0.5ex}
\includegraphics[width=1.0\textwidth,height=0.35\columnwidth]{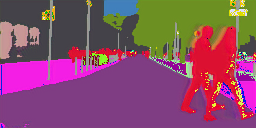}\vspace{0.5ex}
\includegraphics[width=1.0\textwidth,height=0.35\columnwidth]{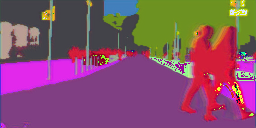}\vspace{0.5ex}
\end{minipage}
}
\caption{Examples of video prediction results with 5 consecutive input frames from the Cityscapes dataset. For each frame, we show (a) the input frames, (b) ground truth frames, (c) results generated by ConvLSTM method and (d) results produced by our networks trained with $\ell_1+\ell_{gdl}$ loss. Different rows show the results of different time steps. The input frames are at $t=1,2,3,4,5$. The ground truth and prediction frames start from the next time step $t=2$ to $t=6$.}
\label{fig:segmentations}
\end{figure*}

\subsection{Video prediction of semantic segmentations}
In addition to the prediction of natural image sequences, we also investigate predicting future semantic segmentations. Semantic segmentation is a simple form of visual understanding, where each pixel has the corresponding label (e.g., vehicle, road, pedestrian, etc.). Making prediction in semantic-level does not consider too much about the detailed textures and edge boundaries. By this way, the prediction model can focus on predicting the motions of different objects. We train the proposed model on the Cityscapes dataset, which has 8K training and 1K testing semantic segmentations. Here we used $\ell_1$ loss function for the video prediction.

We show some prediction results made by our model and ConvLSTM with $\ell_1$ loss function quantitatively in Figure \ref{fig:segmentations}. Our model can make a more accurate prediction than the ConvLSTM. The results generated by ConvLSTM is blurry, and is similar to the input semantic frames instead of the ground truth frames. Though the results by our networks have some noise, the motions of the pedestrian are precisely predicted. The results demonstrate that our model is an effective method to make predictions in semantic segmentation space. 

Table \ref{tab:seg performance} presents the mean pixel SSIM error and PSNR of different methods. The higher PSNR means better prediction performance, and our model trained with the $\ell_1$ loss objective function can achieve better performance. From the SSIM and PSNR in the table, frame prediction generated by our method can make more accurate prediction of future frames.

Overall, our model using optical flow is capable of making accurate video prediction both in RGB-level and semantic-level large-scale scenes.  
\begin{table}[ht]
\begin{center}
\begin{tabular}{l c c}
\toprule
Method & PSNR &SSIM\\
\midrule
ConvLSTM \cite{xingjian2015convolutional}& 18.9462 & 0.8715\\
Our networks & {\textbf{26.3671}} & \textbf{0.9490}\\
\bottomrule
\end{tabular}
\end{center}\caption{Mean pixel SSIM error and PSNR on the Cityscapes dataset for video prediction with two input streams.}
\label{tab:seg performance}
\end{table}

\section{Conclusion}
We present a novel end-to-end learning framework that utilizes the spatial-temporal information of frames that make high quality video prediction. Our model is trained on unlabeled image sequences with an application to diverse and complex urban scenes. The motion of pixels in optical flow is implicitly incorporated in the feature maps. By doing this, our model can predict more accurate future frames both at RGB-level and semantic-level. In addition, our model as a predictive perception part can be easily deployed on the autonomous driving system. There are some other challenging problems for video prediction problems, such as how to make longer prediction. It is difficult to make longer horizon prediction because the uncertainty of the complex and changeable scenes will increase rapidly.
\IEEEtriggeratref{11}
\bibliographystyle{IEEEtran}
\bibliography{vp_MMSP18}

\begin{thebibliography}{10}
\providecommand{\url}[1]{#1}
\csname url@samestyle\endcsname
\providecommand{\newblock}{\relax}
\providecommand{\bibinfo}[2]{#2}
\providecommand{\BIBentrySTDinterwordspacing}{\spaceskip=0pt\relax}
\providecommand{\BIBentryALTinterwordstretchfactor}{4}
\providecommand{\BIBentryALTinterwordspacing}{\spaceskip=\fontdimen2\font plus
\BIBentryALTinterwordstretchfactor\fontdimen3\font minus
  \fontdimen4\font\relax}
\providecommand{\BIBforeignlanguage}[2]{{%
\expandafter\ifx\csname l@#1\endcsname\relax
\typeout{** WARNING: IEEEtran.bst: No hyphenation pattern has been}%
\typeout{** loaded for the language `#1'. Using the pattern for}%
\typeout{** the default language instead.}%
\else
\language=\csname l@#1\endcsname
\fi
#2}}
\providecommand{\BIBdecl}{\relax}
\BIBdecl

\bibitem{finn2016unsupervised}
C.~Finn, I.~Goodfellow, and S.~Levine, ``Unsupervised learning for physical
  interaction through video prediction,'' in \emph{Advances in Neural
  Information Processing Systems}, 2016, pp. 64--72.

\bibitem{krizhevsky2012imagenet}
A.~Krizhevsky, I.~Sutskever, and G.~E. Hinton, ``Imagenet classification with
  deep convolutional neural networks,'' in \emph{Advances in neural information
  processing systems}, 2012, pp. 1097--1105.

\bibitem{srivastava2015unsupervised}
N.~Srivastava, E.~Mansimov, and R.~Salakhudinov, ``Unsupervised learning of
  video representations using lstms,'' in \emph{International Conference on
  Machine Learning}, 2015, pp. 843--852.

\bibitem{xue2016visual}
T.~Xue, J.~Wu, K.~Bouman, and B.~Freeman, ``Visual dynamics: Probabilistic
  future frame synthesis via cross convolutional networks,'' in \emph{Advances
  in Neural Information Processing Systems}, 2016, pp. 91--99.

\bibitem{oh2015action}
J.~Oh, X.~Guo, H.~Lee, R.~L. Lewis, and S.~Singh, ``Action-conditional video
  prediction using deep networks in atari games,'' in \emph{Advances in Neural
  Information Processing Systems}, 2015, pp. 2863--2871.

\bibitem{mathieu2015deep}
M.~Mathieu, C.~Couprie, and Y.~LeCun, ``Deep multi-scale video prediction
  beyond mean square error,'' \emph{arXiv preprint arXiv:1511.05440}, 2015.

\bibitem{lotter2016deep}
W.~Lotter, G.~Kreiman, and D.~Cox, ``Deep predictive coding networks for video
  prediction and unsupervised learning,'' \emph{arXiv preprint
  arXiv:1605.08104}, 2016.

\bibitem{neverova2017predicting}
N.~Neverova, P.~Luc, C.~Couprie, J.~Verbeek, and Y.~LeCun, ``Predicting deeper
  into the future of semantic segmentation,'' \emph{arXiv preprint
  arXiv:1703.07684}, 2017.

\bibitem{ranzato2014video}
M.~Ranzato, A.~Szlam, J.~Bruna, M.~Mathieu, R.~Collobert, and S.~Chopra,
  ``Video (language) modeling: a baseline for generative models of natural
  videos,'' \emph{arXiv preprint arXiv:1412.6604}, 2014.

\bibitem{luo2017unsupervised}
Z.~Luo, B.~Peng, D.-A. Huang, A.~Alahi, and L.~Fei-Fei, ``Unsupervised learning
  of long-term motion dynamics for videos,'' \emph{arXiv preprint
  arXiv:1701.01821}, 2017.

\bibitem{vondrick2016generating}
C.~Vondrick, H.~Pirsiavash, and A.~Torralba, ``Generating videos with scene
  dynamics,'' in \emph{Advances In Neural Information Processing Systems},
  2016, pp. 613--621.

\bibitem{DBLP:journals/corr/WattersTWPBZ17}
N.~Watters, A.~Tacchetti, T.~Weber, R.~Pascanu, P.~Battaglia, and D.~Zoran,
  ``Visual interaction networks,'' \emph{CoRR}, vol. abs/1706.01433, 2017.

\bibitem{horn1981determining}
B.~K. Horn and B.~G. Schunck, ``Determining optical flow,'' \emph{Artificial
  intelligence}, vol.~17, no. 1-3, pp. 185--203, 1981.

\bibitem{xingjian2015convolutional}
S.~Xingjian, Z.~Chen, H.~Wang, D.-Y. Yeung, W.-K. Wong, and W.-c. Woo,
  ``Convolutional lstm network: A machine learning approach for precipitation
  nowcasting,'' in \emph{Advances in neural information processing systems},
  2015, pp. 802--810.

\bibitem{Cordts2016Cityscapes}
M.~Cordts, M.~Omran, S.~Ramos, T.~Scharw{\"a}chter, M.~Enzweiler, R.~Benenson,
  U.~Franke, S.~Roth, and B.~Schiele, ``The cityscapes dataset,'' in \emph{CVPR
  Workshop on The Future of Datasets in Vision}, 2015.

\bibitem{Geiger2012CVPR}
A.~Geiger, P.~Lenz, and R.~Urtasun, ``Are we ready for autonomous driving? the
  kitti vision benchmark suite,'' in \emph{Conference on Computer Vision and
  Pattern Recognition (CVPR)}, 2012.

\bibitem{chollet2015keras}
F.~Chollet \emph{et~al.}, ``Keras,'' \url{https://github.com/fchollet/keras},
  2015.

\bibitem{kingma2014adam}
D.~Kingma and J.~Ba, ``Adam: A method for stochastic optimization,''
  \emph{arXiv preprint arXiv:1412.6980}, 2014.

\bibitem{wang2004image}
Z.~Wang, A.~C. Bovik, H.~R. Sheikh, and E.~P. Simoncelli, ``Image quality
  assessment: from error visibility to structural similarity,'' \emph{IEEE
  transactions on image processing}, vol.~13, no.~4, pp. 600--612, 2004.

\end{thebibliography}
%\bibliography{IEEEabrv,spl}
\end{document}